\documentclass[journal,final]{IEEEtran}
\usepackage{cite}

\usepackage{amsmath,amsfonts}

\usepackage[shortlabels]{enumitem}
\usepackage{bbm}

\usepackage[ruled,linesnumbered]{algorithm2e}
\usepackage{algorithmic}

\usepackage[dvipsnames]{xcolor}

\usepackage{array}

\usepackage{booktabs}

\usepackage{multirow}

\usepackage{pgfplots}
\usepackage{svg}  
\usetikzlibrary{plotmarks}
\usetikzlibrary{patterns}
\usetikzlibrary{arrows.meta}
\usepgfplotslibrary{patchplots}
\usepgfplotslibrary{colorbrewer}
\usepgfplotslibrary{fillbetween}

\usetikzlibrary{external}
\usepackage{pgfplots, pgfplotstable}
\usepgfplotslibrary{groupplots}

\usepackage{filecontents}

\usepackage{textcomp}
\usepackage{stfloats}
\usepackage{url}
\usepackage{verbatim}
\usepackage{graphicx}
\usepackage{xcolor}
\usepackage[font=small]{caption}
\usepackage{subcaption}
\usepackage{multicol}

\usepackage{colortbl}

\usepackage{mathtools}
\newcounter{tableeqn}[table]

\newcounter{tablesubeqn}[tableeqn]

\definecolor{rowgray}{gray}{0.9}

\begin{document}

\title{
Energy-Efficient Federated Learning for AIoT \\ using Clustering Methods
}

\author{Roberto~Pereira,~\IEEEmembership{Member,~IEEE,}
        Fernanda Famá,~\IEEEmembership{Student~Member,~IEEE,}\\
        Charalampos Kalalas,~\IEEEmembership{Member,~IEEE,}
        and~Paolo Dini
\thanks{{All the authors are with the  Sustainable AI Research Unit,  Centre Tecnol\`ogic de Telecomunicacions de Catalunya (CTTC),
Barcelona (Spain).
e-mail: 
\{name.surname\}@cttc.cat}}
\thanks{This work has been partially funded by the grant CHIST-ERA-20-SICT-004 (SONATA) by PCI2021-122043-2A/AEI/10.13039/501100011033
and by the European Union’s Horizon 2020
Marie Skłodowska Curie Innovative Training Network Greenedge (GA. No.
953775).}
}

\markboth{}
{}

\maketitle

\begin{abstract}

While substantial research has been devoted to optimizing model performance, convergence rates, and communication efficiency, the energy implications of federated learning (FL) {within Artificial Intelligence of Things (AIoT) scenarios} are often overlooked in the existing literature. This study examines the energy consumed during the FL process, focusing on three main energy-intensive processes: pre-processing, communication, and local learning, all contributing to the overall energy footprint. We rely on the observation that {device/}client selection is {crucial} 
for speeding up the convergence of model training {in a distributed AIoT setting} 
and propose two clustering-informed methods. These clustering solutions are designed to group {AIoT devices} 
with similar label distributions, resulting in clusters composed of nearly heterogeneous {devices}. 
Hence, our methods alleviate the heterogeneity often encountered in real-world distributed learning applications.  Throughout extensive numerical experimentation, we demonstrate that our clustering strategies typically achieve high convergence rates while maintaining low energy consumption when compared to other recent approaches available in the literature.

\end{abstract}

\begin{IEEEkeywords}
Energy efficiency, Federated learning, AIoT device selection, Clustering.
\end{IEEEkeywords}

\section{Introduction}
\label{sec:intro}

Until recently, most machine learning model training solutions have relied on centralized systems, where datasets were transferred to and processed in powerful data centers. This approach, while effective, often results in privacy concerns and significantly contributes to global carbon emissions. These emissions are attributed not only to the energy required for data transmission and computation, but also to the operational costs associated with maintaining these large servers~\cite{avgerinou2017trends, baliga2010green}. Recent studies, \cite{ahvar2019estimating, guerra2023cost, fl_with_cooperating_devices_iot,  min2019learning, zhu2022energy} to name a few, suggest that transitioning from centralized to distributed computing has the potential to reduce the overall energy consumed to train machine learning solutions.

{In the emerging paradigm of Artificial Intelligence of Things (AIoT),
a large amount of data is generated by distributed devices, which are often resource-constrained in terms of power, storage, and computational capabilities. In this context, traditional centralized learning approaches are often inefficient, as transmitting large volumes of raw data to a central server is not only limited by network bandwidth but also by the devices' battery lifetime~\cite{Imteaj_survey_constraint_fl}. These challenges, coupled with privacy concerns, have led to the growing adoption of distributed learning methods, in which data is pre-processed locally at the edge device.  
}

Particularly, federated learning (FL) has emerged as a promising solution  {to AIoT applications} 
that allows learning and processing to occur close to the data sources rather than relying on cloud computing centers~\cite{zhu2022energy,  Imteaj_survey_constraint_fl, Abdulrahman_motivation_fl_aiot, bonawitz2019towards_fl}. The objective of FL systems is to train a shared global model by leveraging the collective data distributed across various clients/clients without requiring the direct sharing of their data. In this setting, each client trains a local model for a few iterations using its private data and subsequently transmits the locally updated models to a central coordinating server. The server then aggregates these updates to refine the global model, which is broadcast back to all the clients. FL alternates between a local model computation at each device and a round of communication with a (low-energy consuming) central server.

Despite the evident privacy and computational sharing advantages of FL, it also incurs several new challenges. Typical challenges are usually encountered in the communication and convergence rates of the learning process. 
On one hand, having a large number of participating devices is often associated with more data and leads to better and faster convergence of global solutions. On the other hand, the number of participating devices is typically restricted by the communication and energy resources that are available at the central controller\footnote{In this paper, we use the terms controller and server interchangeably.
{Likewise, we also refer to clients and (AIoT) devices interchangeably.}
} and the clients. 
Consequently, a common strategy in FL is to consider only a subset of currently available clients in the learning process. When data is independent and identically distributed (iid) among devices,  randomly pooling a small subset of these devices may lead to accurate global models.

Unfortunately, in real-world applications, it is often the case that data is heterogeneously distributed (non-iid) among the different clients. In such scenarios, the convergence rate of the model has a weak dependency on the number of selected clients and a higher dependency on which client is being selected. In other words, selecting one client or another may lead to a better or worse global model due to variation in local data availability~\cite{li2019convergence, Zili_survey, li2020federated}. 
In such heterogeneous
scenarios, effective client selection plays an important role 
to avoid redundancy and ensure that selected clients contribute meaningfully to the global model.

\subsection{Related Work}

Motivated by the challenges mentioned above, several studies have focused on adaptive {device/client} selection in FL scenarios with heterogeneous data distribution.
A naive approach refers to employing an (almost) unbiased selection strategy by prioritizing clients with a larger number of samples~\cite{li2019convergence, Zili_survey} which is essentially equivalent, in expectation, to the participation of every client after infinitely many communication rounds. Nonetheless,  under heterogeneous label distribution conditions,  this may lead to the redundant client selection scenario, where in a certain round, no sample of a specific label is observed by the local clients, naturally resulting in degraded results.

A more elaborated approach is proposed by the authors in~\cite{cho2020client_power_choice}, wherein clients with higher local losses are prioritized. This approach is particularly interesting, as the central controller benefits from low computational overhead by simply weighting the clients based on their latest reported losses. 
Similarly, the authors in \cite{chen2020optimal} use the norm of the clients' gradient update to adjust the probability of sampling the different devices.
This, however, comes at the cost of neglecting the correlations between the clients and considering their losses independently, which may result in only marginal improvements in performance.

To overcome the limitation of the methods above, the authors in \cite{tang2022fedcor} employ a Gaussian Process (GP) to model the client loss changes in a correlated fashion. This not only addresses the shortcomings of rudimentary loss-based selection, but it also introduces predictive capabilities into the client selection process. 
This approach typically involves two phases:  a GP training and an inference phase. 
In the first phase, the central controller models changes in client losses using a GP; at this stage, clients are randomly sampled. After the GP is trained, it is then used to estimate the loss changes in the next communication rounds (inference phase). 
A sequence of GP training and inferences is repeated until the convergence of the FL algorithm. Although this sequential training may lead to faster convergence rates, in general, it may also require a large amount of energy resources for the GP training phase.

Similarly, DELTA~\cite{wang2024delta} also considers the correlation between clients by leveraging full-client gradient dependence to sample clients with large gradient diversity. This approach aims to exploit learning diversity and accelerate convergence rates. To address the computational challenges of full-client gradient dependence--i.e., requiring all the clients to participate in every communication round--the authors propose estimating the current round's gradient using the gradients from the previous round\footnote{One our proposed solutions, namely \textit{RepClust} (see Sec.~\ref{sec:clustering} for details), draws similar intuitions as the ones in DELTA. Specifically, we sample devices to maximize their diversity. However, unlike active sampling methods that periodically update participation probabilities, \textit{RepClust} performs this selection only once, prior to training, and bases the sampling on devices' data distributions. This simple solution significantly reduces the energy costs of our solution while maintaining acceptable convergence rates.}. 

In this work, we refer to methods that periodically update their probability of sampling as \textit{active sampling} or \textit{active client selection}. As mentioned above, one of the drawbacks of these strategies is the high computational cost associated with updating sampling probabilities, which often scales with the number of communication rounds.

Closely related to what is later presented in this work, there also exist methods that attempt to combine active client selection with clustering solutions based on some predefined criteria. For instance, the authors in \cite{fraboni2021clustered} explore how clustering clients may optimize server-client communications in FL through unbiased client selection. Their results suggest that clustering clients based on factors such as the number of data samples or the similarity of model updates can improve the general convergence of FL models. Indeed, this similarity-based clustering is closely related with prior work on active client selection \cite{cho2020client_power_choice, tang2022fedcor, balakrishnan2022diverse, wang2024delta}, which aims to identify clients that jointly contribute to learning while minimizing variance in the aggregated model.  A downside is that properly identifying similar models may require that all devices to estimate their local gradients and transmit them to the server in every communication round~\cite{song2023fast}; or that the controller keeps a global dataset that will be used to compare the different models in a centralized fashion \cite{morafah2023flis}. Either way, similar to prior client selection works, both approaches incur necessary computation or communication overhead.

In this work, we build upon the methods discussed above and propose two clustering-based solutions designed to: (i) improve convergence speed and (ii) minimize computational and communication overhead, while requiring only a small subset of clients to participate in each communication round. The first objective is achieved by employing group-based sampling strategies, which ensure that the devices participating in each round (collectively) provide a representative description of the entire training dataset. The second objective is addressed by performing a single clustering operation prior to the training phase of the FL algorithm. By adopting this approach, our method significantly reduces computational costs compared to active learning solutions, which require continuous updates to their sampling strategies.

Parallel to client sampling, another widely employed strategy to tackle the data heterogeneity problem involves the generation of synthetic data samples (or features), derived from the data available at the local clients, to be shared with the central controllers. 
This approach aims to enrich the central model training without compromising client data privacy. 
However, this may introduce additional complexities, such as the need for an additional fine-tuning phase~\cite{luo2021no, xu2022fedcorr} or access to publicly available datasets~\cite{learnfromothers}. These additional steps can significantly increase the computational and energy costs of the overall learning process.

\begin{figure}[t!]
\centering
\includegraphics[width=0.9\linewidth]{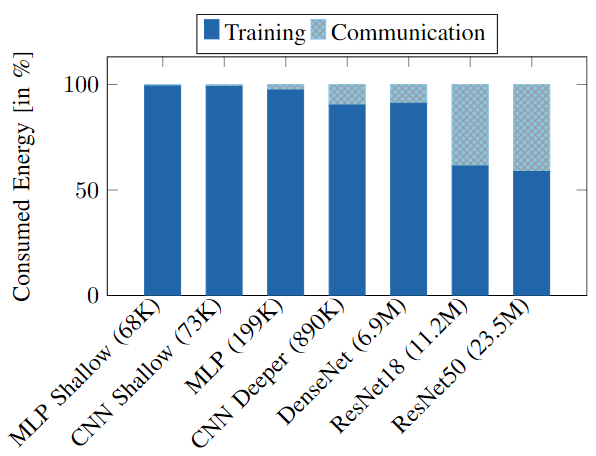}

     \caption{ Percentage of the total energy consumed during training (solid) and communication (striped) stages of a heterogeneous FL process with $100$ clients learning on CIFAR-10 dataset. 
     Random selection of $10$ clients was performed over $100$ communication rounds. 
     The number of learnable parameters of each model is reported inside the parenthesis and alongside the model name on the x-axis.
     }
        \label{fig:all_motivation}
    \vspace{-1\baselineskip}
\end{figure}

\subsection{Contributions}

The majority of the studies above emphasize the energy efficiency of their solutions, primarily by trying to reduce the number of communication rounds between clients and the central controller. 
In this work, we first notice that while a significant amount of energy is consumed during communication, AIoT systems consume an even larger portion of energy during training.
To illustrate this idea, Fig.~\ref{fig:all_motivation} depicts the percentage of total energy consumed during the training (solid area) and communication (dashed area) stages of a heterogeneous FL scenario on the CIFAR-10 dataset~\cite{cifar10dataset}. We follow a similar communication model as in \cite{guerra2023cost, yang2020energy} and consider an IEEE 802.11ax channel access procedure where clients upload their local FL model to the server via frequency domain multiple access (FDMA)\footnote{For each model, we perform $10$ epochs of local training with a batch size of 64 on 10 randomly selected clients (out of 100 available ones) over 100 communication rounds. Details on the implementation of each of the networks together with the source code for all the experiments conducted in this work are available at: {\url{https://github.com/robertomatheuspp/clustering_ee_fl}.
    }
}. We compare the energy consumption (averaged over five different seeds) of different models with increasing complexity (number of learnable parameters).

When examined independently, the raw values (in Wh) of both training and communication energy consumption increase with the complexity of the model. However, a direct comparison, as the one depicted in Fig.~\ref{fig:all_motivation}, reveals that the portion of the energy consumed during training is typically much higher than the portion of the energy spent on communication. This effect is particularly attenuated when considering shallow neural network architectures, which are often required in real-world AIoT applications~\cite{chen2024fedgroup, tang2022fedcor}. 

Specifically, for shallow networks (with less than a million parameters), the energy consumed for communication is almost negligible--typically accounting for less than $10\%$ of the total energy cost--when compared to the training cost.  Training larger models (e.g., ResNet18 and ResNet50) increases the energy consumed by the communication procedure, yet it consistently represents the smallest fraction of the total energy cost. Although these values may vary depending on numerous factors related to the networking (e.g., channel conditions, transmission power, and network capacity),  learning process (e.g., batch size, number of epochs in local training, learning rate, and optimizer), or hardware capabilities, these trends highlight the need to prioritize optimization of the training phase over communication, as it constitutes the primary contributor to energy consumption in FL workflows.

In this context, we argue that to reduce the overall energy cost of distributed learning in AIoT systems, it is not sufficient to solely optimize the client selection mechanism to minimize the number of communication rounds.
Instead, larger energy savings can be achieved by sampling clients based on the quality and distribution of their data. Specifically, in AIoT systems-—where devices are often subject to strict power and computational constraints—-sampling clients with higher data quality and/or representativeness can lead to faster convergence, thereby reducing the number of local training iterations needed and reducing total energy consumption.
In this work, we focus on energy-efficient solutions that come from scalable client selection strategies. While existing client selection techniques often prioritize faster convergence, they typically involve computationally expensive pre-processing steps performed at each communication round
\cite{tang2022fedcor, wang2024delta, chen2020optimal, FL_uncer_correlatedclient, fraboni2021clustered, cho2020client_power_choice}. Consequently, as corroborated by our numerical experiments, the energy saved by reducing a few communication rounds is often compensated (or even exceeded) by the energy expended during the client selection process. 
Notably, a critical yet often overlooked aspect in most of the solutions presented above is the substantial energy involved in actively selecting these clients and/or the pre-processing mechanisms related to it. 

Hence, in this work, we depart from traditional per-round selection strategies and explore a lightweight, one-time client clustering approach.
Rather than performing an active client sampling at every communication round, we propose to cluster clients once,  prior to the start of the FL training process. 
Then, during training, random sampling is performed from each cluster.
As supported by our results, this straightforward approach significantly enhances convergence while reducing the energy costs associated with repeated pre-processing and active client selection. 
This is closely related to our recent work~\cite{fama2024measuring}, which empirically studies how different metrics may benefit the clustering of clients based on their data label distribution. In this current study, we expand upon the analysis conducted in our previous research and propose two clustering mechanisms that improve performance while further alleviating the effects of data heterogeneity.

We can summarize the main contributions of this paper as follows. 
\begin{itemize}
    \item We propose two different methods to cluster clients based on their data distribution. Differently from previous approaches, our proposed clustering solutions can be seen as a pre-processing step that only needs to be executed once before training starts.  
    \item  Our proposed cluster-based client selection solutions focus on reducing the total energy consumed in the FL process rather than focusing only on minimizing the number of communication rounds.
    \item {A comprehensive study is conducted on the energy consumption of various processes involved in training an FL model, including pre-processing on the server side, server-client communication, and local training. }
\end{itemize}

\section{Preliminaries}

Before introducing in Sec.~\ref{sec:clustering} our proposed (energy efficient) clustering-based sampling approach, we first review in this section the main concepts and theory behind heterogeneous FL and active sampling schemes.

Let us consider the scenario where $L$ clients have access to a local dataset, denoted by ${D_1, \ldots, D_L}$. In a typical centralized FL setting, a server learns a global model with weights $w$ by minimizing a weighted global loss function, defined as 
\begin{equation}
    \ell = \sum_{j=1}^L \tilde{\ell}_j(w, D_j),
\end{equation}
being $\tilde{\ell}_j(w, D_j)$ the loss function computed locally by each client.
Generally, this optimization is performed in a distributed manner, where at a given round $t = 1, \ldots, T$, each client $ j = 1, \ldots, L$ initializes their local model $w_j^{(t)}$ with the global one $w^{(t)}$ and later performs $E$ epochs of local training. 
In a typical (centralized) FL process, namely federated averaging  (FedAvg), the server then 
aggregates the local updates and generates the new global model by computing the weighted average of the local updates as
\begin{equation}
    w_{t + 1} = \sum_{j=1}^L \frac{|D_j|}{|D|} w_t^{(j)},
\end{equation}
where $|\cdot|$ denotes the size of the dataset and $|D |= \sum_{j=1}^L|D_j|$. 

. 

In real-world scenarios, the local data $D_j, j = 1, \ldots, L$ available at each device may follow different distributions (e.g., covariate shift, concept drift/shift and label distribution skew~\cite{kairouz2021advances, li2022federated, hsu2019measuring}). 
In this study, {we focus on the last case, namely label heterogeneity,  which pertains to variations in label distribution across} different clients. 
For instance, when clients collect location-dependent data, certain classes may become over- or under-represented in their local datasets, leading to significant imbalance across the system.

Another typical problem in FL concerns the number of participating devices. Specifically, as the number of clients participating in the learning process increases, scalability issues become more pronounced. {A key challenge arises}
in the communication context, where the uplink channels between clients and the central server are usually unstable and rate-constrained. 
In this case, the number of devices that can simultaneously communicate to the server is often limited by the server capacity constraints, e.g., heavy traffic may saturate the network capacity and cause server congestion. 
Thus, several works (see \cite{zecchincommunication_fl, chen2020joint} and references therein) have focused on devising efficient network protocols or optimizing the available network resources. Despite their valuable contributions, these methods are 
often limited by physical constraints such as communication bandwidth, transmission/receiver power, network overload, and uplink communication rates, to name a few. An alternative {approach} 
to mitigate such scalability issues is client selection, where only a subset of clients actively participate in each communication round. In this setting, the central controller 
{determines, prior to each communication round,}
which devices should perform local training while the others remain idle.

\subsection{Active Client Selection via Priority Ranking}

When data is equally distributed among the devices, a simplistic approach, which ensures convergence, consists of randomly sampling $K$ clients with sampling probability $\pi_j, j = 1, \ldots, L$ associated with the portion of data available at each client, i.e., $\pi_j = |D_j| / \sum_{j = 1}^L |D_j|$. Particularly, when considering the homogeneous setting, increasing the number of active devices (e.g., $ K \approx L$) also tends to improve the convergence rate of the FL solution \cite{stich2018local}. However, when considering the more realistic non-iid scenario, recent works  \cite{li2019convergence} have shown that the convergence rate of FedAvg has a weak dependence on $K$. Thus, in practice, one may choose $K/L$ arbitrarily small without strongly affecting the convergence rate. Nonetheless, for $K<<L$, selecting one client or another may lead to different solutions. For instance, it might be the case that in a certain round, no sample of a specific class is observed, which will naturally lead to degrading results.

As mentioned above, a more sophisticated solution involves actively selecting clients based on their historical updates. This motivated, for instance, the power-of-choice (PowerD) client selection approach \cite{cho2020client_power_choice}, which ranks and selects clients according to their local losses. 
Initially, $d \in \mathbb{N}$ clients are sampled (without replacement and for $d > K$) based on the probability  $\pi_j, j = 1, \ldots, L$. In a second stage, these clients receive the global model and estimate their local losses, which are then sent back to the server. Finally, the server selects the $K$ clients with the highest losses for local training.
The {rationale} is that only the clients with high losses should perform local training. 
One of the primary observations motivating PowerD is that the update of a single client will affect the global model; {thus,} ranking and choosing clients based on their losses may benefit the convergence of the global model. 
Unfortunately, while intuitive, this method has shown \cite{chen2020convergence, chen2022optimal_sampling, tang2022fedcor} only marginal improvements over random sampling in complex tasks (e.g., CIFAR-10 dataset).

To further improve the overall convergence of the system, more recent approaches have sought to rank clients based on their joint contribution to the global model. Specifically, the authors in \cite{chen2022optimal_sampling} introduce an \textit{important update} client selection solution where clients are jointly ranked based on their norm updates. 
Similar to the above, the goal is to select, at each communication round, a subset of clients that will contribute with the most informative updates possible. 
However, we argue that client selection should also consider the correlation among clients, as there may exist clients that send very similar (or very different) updates. From a network communication perspective, sending the norms is relatively inexpensive.  Nonetheless, obtaining the norm updates requires clients to perform local training, which may incur high energy costs.  

A similar approach, which will be considered as a baseline in this work, but that strongly alleviates the need for constant local updates is to model the change in the client losses using a GP \cite{tang2022fedcor}. One of the advantages of considering a GP is that we no longer have to obtain the norm updates from the clients. Instead, the loss changes can be directly predicted from the GP. Finally, clients can be selected such that they minimize the posterior expectation of the overall loss conditioned on their loss changes. 
One drawback of this approach is that all clients must periodically update their model and send it to the server to train the GP. These extra training steps heavily increase the overall energy cost of such methods.

From our literature review, it {becomes apparent} that most of the efforts have focused on increasing convergence speed while minimizing communication overhead. We claim here that this approach marginally considers the total energy consumed in the process (including both training and communication energy, as per Sec.~\ref{sec:intro}) and, in particular, almost disregards the cost of local training on the client side.
In what follows, we consider an alternative approach that takes into account the total energy consumed in the FL process. 
Specifically, rather than 
{periodically performing} an active selection/{ranking} of clients, 
we {conduct} a light pre-processing step that clusters clients based on their label distribution\footnote{
We {acknowledge} that this {approach} may raise privacy {concerns}, which we leave  for future work. Instead, this work advocates for 
the importance of data similarity-based clustering for achieving energy-efficient FL.}. After being clustered, these clients are then sampled such that the selected ones jointly contribute to the learning solution. Specifically, we aim to minimize the selection of clients that provide redundant contributions to the global model (e.g., clients with highly similar data distributions). 
In the subsequent sections,
we detail the clustering mechanisms that we consider in this work together with their associated sampling mechanisms.

\section{Clustering-Informed Client Selection Mechanisms}
\label{sec:clustering}

\begin{figure}[t!]
    \centering
    \includegraphics[width=1\linewidth]{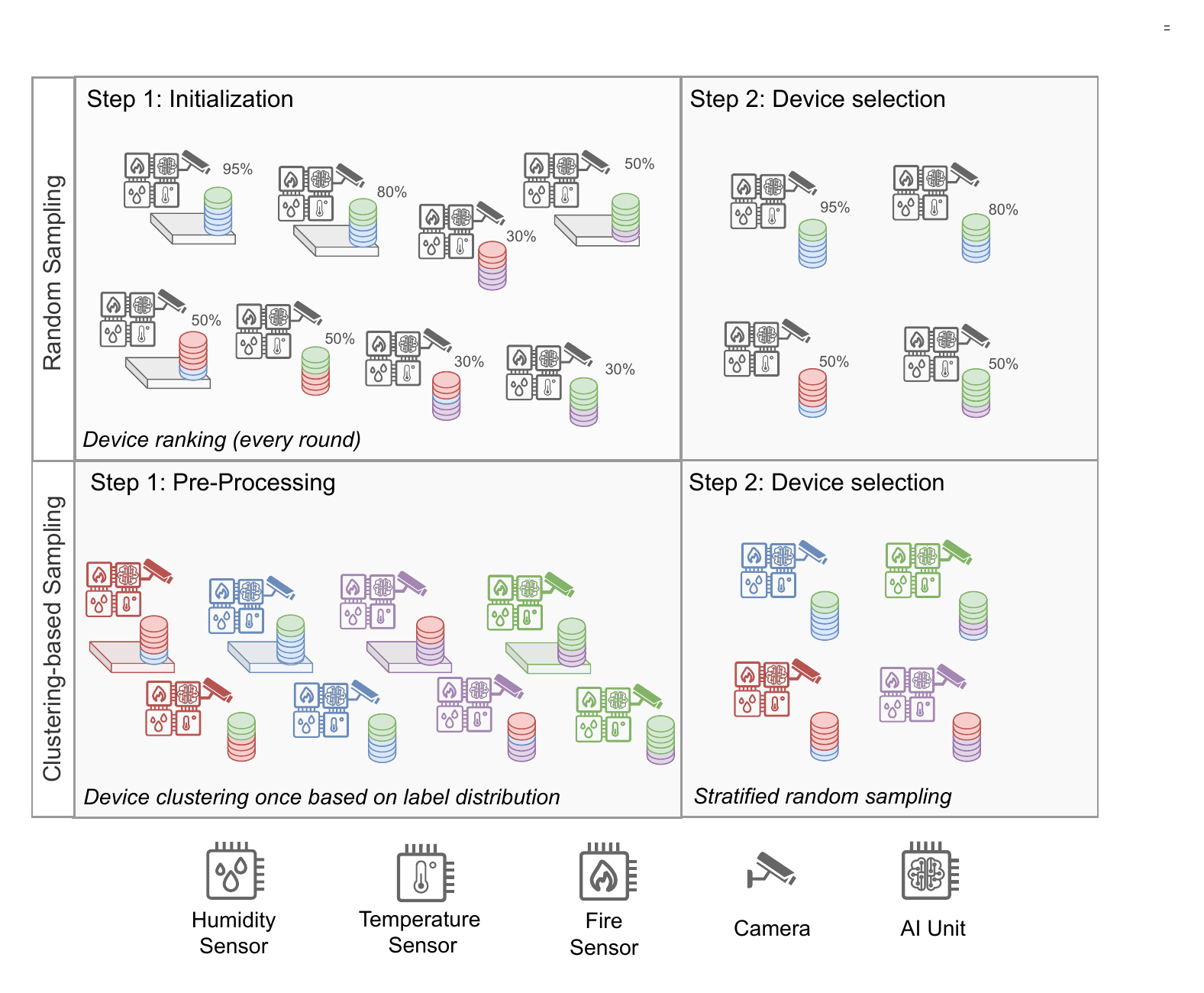}
    \caption{
    Comparison {between} traditional 
    {device} 
    random sampling (first row) {and the} proposed 
    {device}  sampling (second row).
    The circular disks next to {each} 
    {device}  represent their data and label distribution (i.e., each color denotes a different label).
    In random sampling, the numbers next to the data represent the ranking of the 
    {device}. 
    In the second row, clients with the same colors {exhibit} similar label distributions. 
    }
    \label{fig:illusrative_clientselectoin}
\end{figure}

In heterogeneous settings, the quality and diversity of data available across different clients can vary significantly. In real-world scenarios, data is often location-dependent, {implying} 
that clients in nearby areas may share similar data {characteristics}~\cite{li2022federated}. 
In such scenarios, the controller must {select clients judiciously}  
\textit{i}) to ensure that the chosen participants are representative of the global population and {\textit{ii}) prevent bias} in the global model towards the local data of those clients.

To {tackle} this challenge, we will explore two alternatives to simple random sampling. In both {approaches}, we first perform a clustering step that {remains} fixed during the whole training\footnote{This procedure is conceived under the assumption that data do not suffer from any temporal drift and training is performed under already {acquired} local training data.} and {subsequently}, at each communication round, we sample clients in a way that ensures the {group}
of clients participating in each round is (almost) homogeneous. 
Fig.~\ref{fig:illusrative_clientselectoin} compares random and clustering-informed client selection. The different colors of the data (barrels next to the client) represent the label distribution within each local dataset. Notice (in step 2) that performing random sampling may result in very few samples from the purple label, whereas clustering-informed sampling leads to a more balanced {representation}.

More formally, let us consider an FL setup where the downstream task consists of $M$ classes divided among $L$ clients, each of which is associated with a label distribution  $\mathbf{p}_\ell = [p_{\ell, 1}, \ldots, p_{\ell, M}], \, \ell = 1,\ldots, L$. Let us also define the difference between two clients as $d(i,j) \geq 0, \, i,j  = 1,\ldots, L$. We {aim} to form $G$ disjoint partitions  $\{\mathcal{C}_1, \ldots, \mathcal{C}_G\}$, where $\mathcal{C}_g, g = 1, \ldots, G$ indexes the clients associated with the $g$th partition.
At each round, the controller uses these partitions to select clients such that their data is as similar as possible and {collectively represent} 
the majority of the data space. 
Hence, in what follows, we will {describe}
two {potential} clustering mechanisms that 
{aim to}
\textit{i}) maximize similarity among clients within the same cluster; and \textit{ii}) maximize diversity within each group.

In the first clustering mechanism, the controller is interested in clustering together clients that have similar label distributions. 
{Subsequently, in each round,}
only one element per cluster is selected to perform local training while the others remain idle. The drawback of this traditional clustering mechanism is {the ambiguity regarding}
which clients to select from each cluster. Hence, we also consider grouping together elements that maximize the diversity {within} each group. In this scenario, {our aim is} to cluster together clients such that their combined data collectively represent the majority of the data space in an (almost) iid fashion. In this maximum diversity setting, in each round, one cluster is selected, and all clients associated with this cluster {engage in local training}.

\subsection{Clustering Similar Clients Together}

An alternative to {simple random sampling},
commonly found in the 
Neyman\textquotesingle s~sampling theory~\cite{neyman1992two_stratified}, is the so-called stratified random sampling. In this method, the population is first divided into distinct, almost homogeneous subgroups known as strata. Once the population is stratified, a simple random sample is drawn from each subgroup. In the FL context, this translates into first grouping clients such that the distance $d(\mathbf{p}_j, \mathbf{p}_k)$ is minimized for all pairs of clients $j  \neq k \in \mathcal{C}_g$ and all possible partitions $g = 1, \dots, G$. Hereafter, we will 
{refer to} this clustering approach {as} \textit{similarity clustering}, 
{since it aims to combine clients with similar label distributions}.

Intuitively, the goal is to maximize the similarity of the clients forming the same group. This can typically be obtained using classical clustering approaches, for instance, by employing k-Means 
with symmetrized Kullback-Leibler (KL) divergence as a metric. Arguably, the choice of the clustering algorithm and its metric may depend on the actual scenario. In this study, we consider the symmetrized KL-divergence due to its statistical interpretation, and k-Means
{which has shown to yield the best results in our previous experiments \cite{fama2024measuring}.}
At each communication round, the central controller needs to select  $K$ clients to participate in the training. When considering the similarity clustering-based solutions, if $G \geq K$, we randomly select one client of $K$ randomly selected groups. Otherwise, if $K > G$, we randomly select at least $K/G$ clients per group to participate in the training. For a fair comparison with the other baselines, during our experiments, we always ensure that we select exactly $K$ clients to participate in each FL round. Our intuition suggests that it is always beneficial to have $G$ smaller or close to $K$. The goal here is to cluster together clients {with} 
similar data and {minimize} 
redundancy, so that when choosing clients we are likely to {select} the ones that contribute with different portions of the distributed data.

\subsection{Maximizing Group Diversity}

One of the {issues} with the above approach is that one needs to decide how 
to draw elements from the different clusters at each communication round, e.g., using random selection or round-robin scheduling. Nonetheless, there is no guarantee that clients selected from different clusters follow any 
requirement. For instance, it might be the case that 
selected clients
{jointly do not have samples from one or multiple classes.}
To avoid such undesired behavior, inspired by the maximally diverse group problem and the pilot assignment problem in the wireless communication context \cite{mohebi2022repulsive, lee2020deep}, we propose to cluster clients such that each group is as diverse as possible. Additionally, different clusters should be as similar as possible. In this scenario, instead of choosing one random element from each cluster, one selects a whole partition for each communication round. By doing so, we ensure that at each communication round, {we minimize training over (possibly) redundant data by maintaining diversity among the selected clients.}
{We will refer to this method as} \textit{repulsive} clustering in our numerical experiments.

Formally, we aim to solve the following multi-objective optimization problem
\begin{equation}
\begin{aligned}
\max_{\{\mathcal{C}_1, \ldots, \mathcal{C}_G\}} \quad &
\frac{1}{G}
\sum_{g = 1}^G
\frac{1}{|\mathcal{C}_g|(|\mathcal{C}_g| - 1)}
\sum_{i,j \in \mathcal{C}_g} d(\mathbf{p}_i,\mathbf{p}_j),
\\
\min_{\{\mathcal{C}_1, \ldots, \mathcal{C}_G\}} \quad & \frac{1}{G(G - 1)} \sum_{g = 1}^G \sum_{g^\prime = g + 1}^G
d\left( \mathbf{p}^{(g)}, \mathbf{p}^{(g^\prime)}\right), 
\\
\textrm{s.t.} \quad &  |\mathcal{C}_g| = |\mathcal{C}_{g^\prime}|, 
\quad \text{ for } g \neq g^\prime,  g, g^\prime = 1,\ldots, G,
\end{aligned}
\label{eq:max_diversity_optmization}
\end{equation}
where we have defined the label distribution associated to the $g$th group as $p^{(g)} = \frac{1}{|\mathcal{C}_g|}\sum_{j \in \mathcal{C}_g} \mathbf{p}_j, g = 1, \ldots, G.$
The first objective function promotes diversity within the clusters, while the second one encourages similarity among different clusters. 
The rationale is that different clusters should have similar data distribution, so that there is little difference between the overall data observed by two distinct communication rounds. 
The constraint forces clusters to {be of the same size.}
Removing this constraint leads to trivial solutions where one cluster has $L - G + 1$ elements and the {remaining} 
$G - 1$ clusters {each} have one element. 
We could also {achieve this}
with a tunable parameter. Moreover, from an energy perspective, it {is reasonable} 
that different communication rounds {should} consume similar {amounts of} energy. 

Unfortunately, this is an NP-hard problem \cite{lee2020deep, ravi1994heuristic}, {but with certain} relaxations, it can be solved {via} integer programming. Here, we adapt the heuristic algorithm presented in \cite{mohebi2022repulsive} to find a feasible ({though} not necessarily optimal) solution to the repulsive clustering problem. The solution {employed} in this work is {detailed} 
in Algorithm~\ref{alg:heuristic_max_diversity}.

\begin{algorithm}[t]
\KwIn{
    The label distribution $p_j, j = 1, \ldots, L$ of every client,  the desired number of groups $G$ and the maximum number of search interactions $S$. \\
}
\KwOut{Partitions $\mathcal{C}_g, g = 1, \ldots, G$.}
Randomly initialize  groups $\mathcal{C}_1, \ldots, \mathcal{C}_G$ \\
\While{diversity (\ref{eq:max_diversity_optmization}) improves}{
\For{$ g = 1, \ldots, G$ }{
    \textbackslash \textbackslash Pairs with the smallest distance in each cluster
    \\
    $i_g, j_g \leftarrow \arg\min d(\mathbf{p}_i, \mathbf{p}_j)$, where $i \neq j \in \mathcal{C}_g$
} 
    \textbackslash \textbackslash Sort clusters based on smallest intra-cluster pairwise distances
    \\
    $\tilde{g}_1,\ldots, \tilde{g}_G \leftarrow  \arg \min d(\mathbf{p}_{i_g}, \mathbf{p}_{j_g})$
    \\
    \textbackslash \textbackslash Iterate over top-$T$ clusters
    \\
\For{$k  = 1, \ldots S$}{
    \For{$\ell  = k + 1  \ldots  S$}{
        \If{swapping element $i_k$ from cluster $\mathcal{C}_k$ with 
        element $i_\ell$ from cluster $\mathcal{C}_\ell$
        increases diversity in  (\ref{eq:max_diversity_optmization})}{
        Swap clusters of $i_k \leftrightarrow i_\ell$ in $\mathcal{C}_k$ and $\mathcal{C}_\ell$
        } 
    }
}
}
\KwRet{Partitions $\mathcal{C}_1, \ldots, \mathcal{C}_G$}
\caption{Heuristic solution to (\ref{eq:max_diversity_optmization})}
\label{alg:heuristic_max_diversity}
\end{algorithm}

\subsection{Scalability of Clustering Solutions}
\label{sec:scalability}
Scalability is an important aspect of AIoT network, requiring efficient solutions to both small and large-scale settings.  
We emphasize that the clustering algorithms are executed only once before the training phase, incurring negligible processing costs compared to the overall FL training process. Nevertheless, to further highlight the computational costs, we also analyze the computational complexity of the proposed clustering solutions.
For similarity-based clustering, we observe that the typical solution (using k-Means) has time complexity $\mathcal{O}(KG)$, with $K$ and $G$ being the total number of clients and groups, respectively. 

Analyzing the time complexity of the repulsive clustering solution described in Algorithm~\ref{alg:heuristic_max_diversity} is slightly more complex. Specifically, finding the smallest pairwise distance in each cluster (lines 5) requires pairwise comparisons of all the elements, leading to time complexity $\mathcal{O}(\frac{1}{2} (K/G)^2)$. Similarly, sorting the clusters (line 8) incurs a complexity of $\mathcal{O}(G\log G)$, while the swapping mechanism is $\mathcal{O}(S^2)$. During our experiments, we observed that for repulsive clustering, the ratio $K/G$ is typically small, as a result the overall time complexity of the repulsive clustering algorithm can be approximated by
$$
\mathcal{O}\left( I \left( \frac{1}{2}\frac{K}{G^2} + G\log G + S^2\right)\right) \approx \mathcal{O}\left( I \left( G\log G + S^2\right)\right),
$$
where $I > 0$ represents the number of iterations before convergence. 

These theoretical trends are further validated in Appendix~\ref{appendix:scalability_clustering}, where we present numerical results demonstrating the energy consumption using both algorithms in small and large-scale settings. These results corroborate that the clustering solutions effectively balance computational efficiency while maintaining scalability in large AIoT networks.

\subsection{Privacy Concerns}

Directly sharing the label distribution of clients with a centralized server naturally raises privacy concerns, as it may expose sensitive information about the client's local data. While we acknowledge this issue, we argue that this information can also be inferred from the final layer of local models during training~\cite{ramakrishna2022inferring, yang2023neural}. Specifically, the softmax outputs or logits of the last layer tend to reflect the label distribution implicitly, making it possible for an attacker to estimate class proportions even without explicit access to the true distribution. This suggests that, in practice, an attacker with access to model updates could already infer this information during the training process, regardless of whether the label distribution is explicitly shared. In fact, as explained above, this information has also been previously explored to perform client selection \cite{luo2021no}, but has led to only marginal performance improvement. 

To further address privacy concerns, in Appendix~\ref{appendix:clustering_noise_label}, we have also evaluated the impact of applying local differential privacy mechanism to the label distribution. Our initial findings suggest that even under high levels of privacy protection, the clustering-based selection remains effective, achieving strong accuracy while maintaining energy efficiency.
In this manuscript, we soften the privacy requirement on the label distribution and showcase how correctly selecting clients based on their local data is crucial to improve the overall energy efficiency of traditional FL frameworks. 
While this approach raises privacy concerns, we emphasize that similar information could still be inferred by an attacker through alternative mechanisms. Moreover,  we also highlight that clustering performance remains robust even with strong privacy masking via differential privacy. Overall, this work can serve as a baseline for future studies exploring clustering solutions under stricter privacy constraints.

\section{Energy Consumption of Federated Learning Process} 
\label{sec:energycons}

{During} the learning {phase}, several {components} contribute to overall energy consumption, which can generally be categorized into three main areas:
computations at the server, local training, and communication. 
{Accordingly,} the total energy associated {with} $k$th client, $k = 1, \dots, L$, can be decomposed as
$$
E^{(k)} = E_{\mathrm{pre}}^{(k)} + E_{\mathrm{train}}^{(k)} 
+ E_{\mathrm{comm}}^{(k)}.
$$
When considering random sampling, computations at the server, $E_{\mathrm{pre}} =  \sum_{k = 1}^{K} E_{\mathrm{pre}}^{(k)}$, {are} usually 
{limited} to model aggregation and are often {negligible}
compared to other energy costs (e.g., local training and communication). However, when {employing}
more sophisticated approaches that perform ranking, active client selection (e.g., \textit{FedCor} and \textit{Power}-$d$), or clustering of clients, the energy cost of 
{these} operations may no longer be {negligible}.
In the subsequent sections, we detail the procedure considered to measure the total hardware and communication energy consumption.

\subsection{Hardware Energy Consumption: CPU, GPU and Memory}
\label{sec:energycons:hardware}

We use Codecarbon Python library \cite{codecarbon, bannour2021evaluating} to track hardware-related energy consumption ($E_\text{pre}$ and $E_\text{train}$) which allows us to measure energy consumption associated with the CPU, GPU, and memory. 
Codecarbon measures the energy consumption of the CPU and GPU by 
periodically tracking (every $\Delta t = 15$ seconds) the power supply $P_\text{unit}(t_i), t_i = 1, \ldots, N$ of each processing unit. CPU energy consumption is tracked using the Intel Running Average Power Limit (RAPL) system management interface, whereas GPU energy consumption is monitored using the \textit{pynvml} Python library~\cite{bouza2023estimate}.
The total energy consumed by the processing units,
$$
E^{(k)}_\text{units} = \sum_{i=1}^N \left(P_\text{CPU}(t_i) + P_\text{GPU}(t_i)\right) \Delta t,
$$
depends on the duration of the FL training process ($N  \Delta t$) and the usage of the resources at each time $t_i = 1, \ldots, N$. 

Codecarbon also tracks memory usage, for which we use the default ``machine'' mode, considering approximately
$37.5\times 10^{-2}$ W/Gb of memory used~\cite{maevsky2017evaluating_memory}. The process is similar to the above and periodically measures the resource consumption before converting it into energy consumption. Formally, 
$$
E^{(k)}_\text{memory} = 
(37.5\times 10^{-2})\sum_{i=1}^N \Omega_\text{memory}(t_i) \Delta t,
$$
where $\Omega_\text{memory}(t_i)$ denotes the amount of memory (in Gb) consumed at each time $t_i = 1, \ldots, N$.

Finally, {this approach allows us to estimate the accumulated energy}
consumed during the pre-processing and training stages from the energy consumed by the processing units (CPU and GPU) and the memory, {as}
$$
E_\text{pre} + 
E_\text{train}
=
\sum_{k=1}^K E^{(k)}_\text{units} + E^{(k)}_\text{memory}.
$$
This comprehensive tracking enables the comparison of the hardware-related energy consumption required by different client selection solutions.

\subsection{Over the Air communication}
\label{sec:energycons:communication}

After local training, we assume that clients upload their local FL model to the server via  IEEE 802.11ax wireless links \cite{guerra2023cost, bellalta2016ieee}. Specifically, we consider the scenario where link conditions {remain} 
constant throughout the entire procedure. In {this} 
case, the total energy consumed during communication can be estimated as 
\begin{align*}
E_\text{comm}^{(k)} &=
E_\text{uplink}^{(k)} + E_\text{downlink}^{(k)} = 
\\
&=\sum_{t = 1}^{T} \Delta_T(|w_t|)\Biggl( 
\mathbbm{1}_\text{up}(t,k)  P_\text{up} +
\mathbbm{1}_\text{down}(t,k)  P_\text{down}
\Biggr),
\end{align*}
where $ \mathbbm{1}_\text{down}(t,k),  \mathbbm{1}_\text{up}(t,k)$ {are} the indicator functions {denoting} whether the $k$th client is participating in the $t$th communication round in the downlink and uplink, respectively. Similarly, $P_\text{down} = 20$dBm and $P_\text{up}=9$dBm are the uplink and downlink transmission powers,
Finally, $\Delta_T(|w_t|)$ is a function that determines {the}
duration for transmitting model weights $w_t$ at a specific time $t=1,\ldots, T$. 
This function depends only on the number of trainable parameters of the network (denoted by $|w_t|$) and several control data (see~\cite{guerra2023cost, bellalta2016ieee} for a detailed model).

{We note that more sophisticated wireless communication protocols could yield more efficient solutions. Nonetheless, the clustering-based client selection mechanisms presented in this work are agnostic to the transmission medium and, therefore, can be applied to various transmission models, conditions, and protocols.}


\section{Numerical Evaluation}
\label{sec:results}

\begin{figure*}
    \centering
    \includegraphics[width=0.95\linewidth]{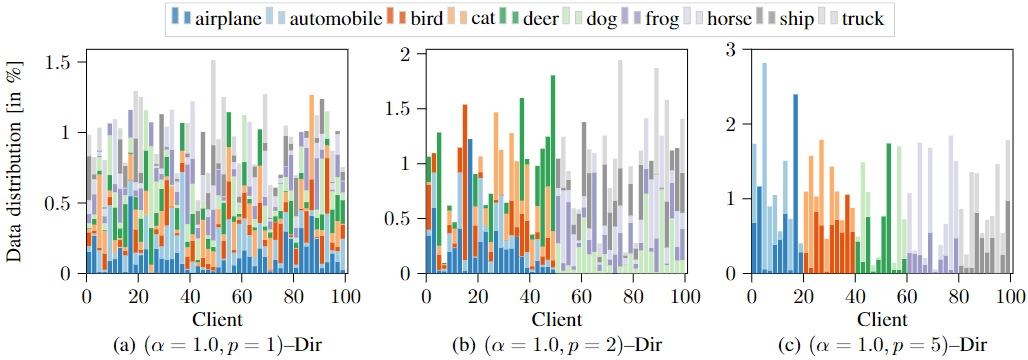}
    \centering 
     \caption{Illustration of different data partition schemes for CIFAR-10 datasets for fixed $\alpha = 1.0$  and varying $\rho$. The higher the $\rho$, the higher the data heterogeneity. 
     }
\label{fig:example_label_distribution}
\end{figure*}

In this section, we evaluate the learning impact of clustering AIoT devices (hereafter referred to as clients) prior to starting the FL training. We conduct experiments on three datasets, namely F-MNIST~\cite{xiao2017fmnist_dataset}, CIFAR-10~\cite{cifar10dataset}, and CIFAR-100, employing a distinct neural network architecture for each, as detailed below\footnote{Further details on model architectures, hyperparameter tuning, and data partitioning (train/test) are provided in Appendix~\ref{sec:nn_details}.}:
\begin{itemize}
    \item F-MNIST: A shallow three-layer feed-forward neural network model (MLP) with of $52,500$ trainable parameters;
    \item CIFAR--10: A convolutional neural network consisting of three convolutional layers followed by two fully connected layers~\cite{tang2022fedcor} totaling $73,418$ trainable parameters;
    \item CIFAR-100: We consider a ResNet18 and train it from scratch, totaling $11,22$M trainable parameters;
\end{itemize}
We consider an FL setup with $L=100$ clients. For reference, the test accuracy of the above models using vanilla FedAvg with full client participation over $T=500$ communication rounds is as follows: F-MNIST ($86.11\%$), CIFAR-10 ($62.43\%$), and CIFAR-100 ($41.00\%$).

Finally, to ease comparison and reproducibility, we report all energy consumption values relative to the total energy consumed over $500$ rounds of communication in a heterogeneous FL process with random client selection of $10$ (out of $100$) clients. 
We measure energy consumption during pre-processing, clustering, and training using the Codecarbon Python library~\cite{codecarbon, bannour2021evaluating} as described in Sec.~\ref{sec:energycons:hardware}, while communication energy is computed as detailed in Sec.~\ref{sec:energycons:communication}. For example, randomly selecting $10$ clients, the total energy consumed is approximately $80.58$ Wh for F-MNIST, $70.10$ Wh for CIFAR-10, and $658.22$ Wh for CIFAR-100.
By setting this standard benchmark, we can clearly assess the energy efficiency of different methods under varying data distribution scenarios, while also allowing for reproducibility using hardware configurations different {from} those considered in our simulations.

\subsection{Experiment Setup}
\label{sec:results:experiment_setup}

In our experiments, for each dataset, we evaluate {various} heterogeneous scenarios and data partitions across $L=100$ clients, as illustrated in Fig.~\ref{fig:example_label_distribution}.
Moreover, we select $K=10$ clients in all client selection mechanisms. {At each client, we perform $10$ rounds of local training\footnote{{We note that, for the F-MNIST dataset, selecting a smaller number of local epochs may improve the convergence of the global model. This may be attributed to potential overfitting to the local data in highly heterogeneous scenarios (e.g., $\rho=5$ partitions).}}. }
We simulate varying levels of data heterogeneity across clients by sampling the ratio of data associated with each label from a Dirichlet distribution. The distribution is controlled using a concentration parameter $\alpha$, which, unless otherwise specified, 
{is set to} $\alpha = 1.0$.
Moreover, 
we are interested in the scenario where data is location-dependent. To simulate this scenario, we first divide the original dataset into $\rho$ partitions{, with} each partition containing all the samples of one or several classes. 
For exposition, we will often consider $\rho=1,2,5$. In the case where $\rho = 1$, we have the traditional scenario where all the clients may contain samples from all classes.

For $\rho = 2$, we form two groups of clients: the first group ($k = 1, \ldots K/2$) has access only to samples associated with the first $M/2 = 5$ labels, while the second group $(\{K/2 + 1, \ldots K\})$ can access only samples corresponding to the remaining five labels.
We denote by $(\alpha, \rho)$--Dir the data that is divided into $\rho$ partitions and has a concentration parameter $\alpha$ within each of these partitions. 
An example is depicted in Figs.~\ref{fig:example_label_distribution}(b) and (c) for $\rho =2$ and $\rho=3$ with $\alpha=1.0$, respectively.  Notice that even for a fixed $\alpha$,  increasing $\rho$ will also impact data distribution -- higher $\rho$ will lead to more heterogeneous scenarios. Finally, we also emphasize that having data location-dependent is typically harmful to the learning process, as the global model may suffer from catastrophic forgetting depending on the sequence of clients considered.

In all the experiments, we use FedAvg as the FL optimizer and report the average results based on five random seeds over $T=500$ communication rounds. 
We compare our proposed pre-processing strategies, namely similar (\textit{SimClust}) and repulsive clustering (\textit{RepClust}), with two other client selection mechanisms\footnote{A detailed explanation on the parameter of each of these methods is available in Appendix~\ref{sec:nn_details} and in the publicly available source code of this work.}: random selection (Rand) 
and \textit{FedCor} \cite{tang2022fedcor}. The first represents a simple client selection mechanism {that results in low} 
energy consumption but at the cost of lower 
accuracy rates in highly heterogeneous scenarios.
 \textit{FedCor} involves two phases: a GP training phase and an inference phase. In the first phase, it models changes in client losses using a GP, {with clients randomly sampled}.
 After the GP is trained, it is then used to estimate the loss changes in the next communication rounds (inference phase). {This} sequence of GP training and inferences is repeated until the convergence of the FL algorithm. 
Although this may lead to faster convergence rates, \textit{FedCor} may also require {a significant} 
amount of energy resources for the GP training phase. Unless otherwise specified, when considering our proposed clustering solution, we select the number of clusters $G \in \{2,5,10,20,25,50\}$ that minimizes the overall energy costs.

Additionally, as discussed {in Sec. \ref{sec:energycons}}, the energy consumed by the FL procedure can be divided into three {components}: the energy consumed for processing at the central controller, local training, and communication. The consumption in each of these {components} 
is significantly influenced by the choice of client selection strategy, the data distribution, and the number of selected clients. 
Therefore, in what follows, {we} will delve into the convergence behavior of the different models and their associated energy consumption, providing insights into the trade-offs between model performance and energy efficiency.


\subsection{Energy Profiles and Convergence Rates}

\begin{figure*}[t!]
\centering
\includegraphics[width=0.98\linewidth]{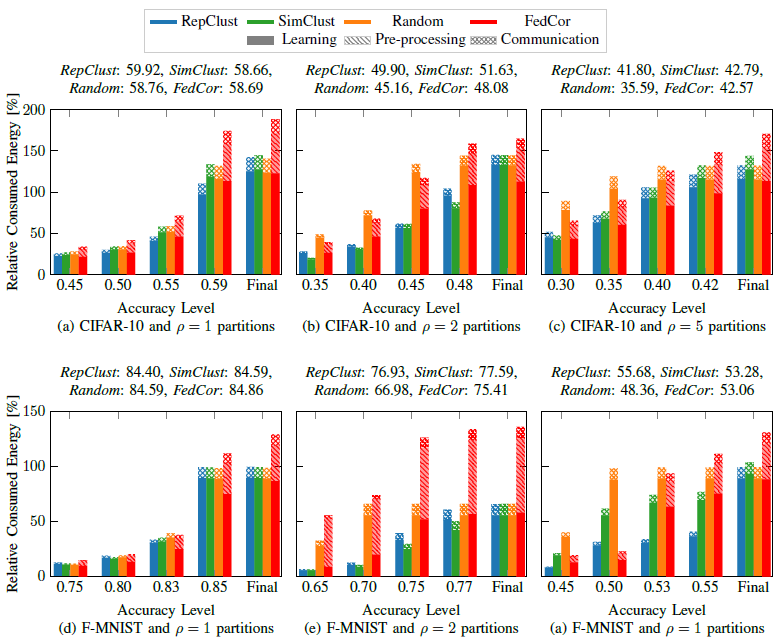}

     \caption{ Energy profiles of client selection mechanisms ({represented by} different colors) divided into learning (solid color at the bottom), pre-processing (diagonal pattern in the middle), and communication (cross pattern at the top). Energy values are reported as relative to the total energy consumed in the FL process with a random selection of $10$ out of $100$ clients for CIFAR-10  (a)-(c) and F-MNIST (d)-(f). Final accuracies are reported inline above each scenario.  
 }
  \label{fig:pick10users_energy}
    \vspace{-1\baselineskip}
\end{figure*}

We start by {examining} 
the energy consumption and convergence rates of the different client selection strategies 
on the CIFAR-10 and F-MNIST datasets
across different heterogeneous scenarios. Specifically, to mitigate the effect of randomness in the results, we focus on the total energy required to achieve a specific accuracy over at least $20$ consecutive epochs\footnote{Our empirical {findings indicate} 
that {varying the} 
number of epochs may lead to different evaluation points, but the {overall} 
conclusion remains consistent to the ones presented in this {study}.}. 
The {objective} 
is to {analyze} 
the energy consumption after the model has already reached some stability {at a desired accuracy level.} 
Fig.~\ref{fig:pick10users_energy} presents accuracy as a function of the total energy consumption, 
distinguishing the amount of energy consumed during training (solid colors), pre-processing (diagonal lines pattern), and communication (cross pattern) for each selection mechanism across CIFAR-10 and F-MNIST datasets. We also report the total (relative) energy obtained after the FL training process is complete.

Regardless of the client selection mechanism, the energy consumed during (local) training {(Fig.~\ref{fig:pick10users_energy})} is {shown to be} always considerably {higher compared to} 
the energy consumed in any of the other stages. 
Specifically,  when employing random selection or either of the clustering-based solutions, the energy {expended} 
during the pre-processing stage becomes almost negligible. 
In other words, performing these operations (i.e., clustering and sampling) incurs minimal energy cost for the FL process. This is expected as clustering is only performed once before training and the computational cost of random sampling grows linearly with the {number} 
of selected samples. 
Conversely, when employing \textit{FedCor},  the amount of energy dedicated {to} pre-processing becomes {notably} 
large and quickly increases with the number of communication rounds. We argue that this increase can typically be associated {with} the ranking of clients and the training of the GP.

In terms of energy required to consistently achieve a specific accuracy level, 
\textit{SimClust} and \textit{RepClust} typically {demand} 
less amount of energy than \textit{FedCor} and random sampling.  Specifically, 
{energy savings during training with \textit{FedCor} are often offset by higher pre-processing costs}
-- e.g., to reach accuracies of $0.48$ and $0.42$ as evidenced in Fig.~\ref{fig:pick10users_energy}(b) and (c), respectively. Conversely, when employing random selection, the energy saved during pre-processing is often consumed in training as random sampling tends to require more communication rounds (i.e., it has a lower convergence rate). 
{Clustering methods appear to strike a balance between these approaches, significantly alleviating the need for pre-processing while maintaining efficient local training.}

{In general}, clustering {clients} 
according to their label distribution allows {for} achieving the desired levels of accuracy with fewer communication rounds, thereby reducing the energy spent on communication. 
Combined with light{weight} pre-processing algorithms, {this approach} helps reduce the total energy costs while speeding up convergence.  
The {synergy between} 
 energy efficiency and lightweight processing is of utmost importance in edge computing, {especially} with resource-constrained devices, and is essential for developing sustainable solutions by {optimizing}
 the available resources.

\begin{figure}
    \centering
    \includegraphics[width=0.98\linewidth]{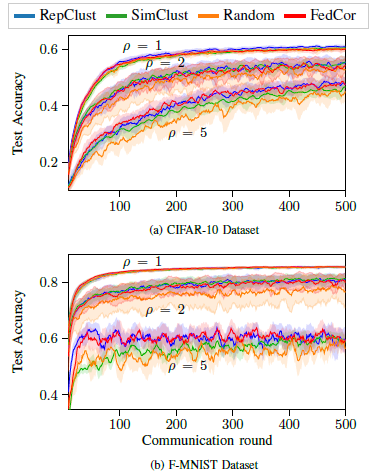}
    
    \caption{Accuracy plots for (a) CIFAR and (b) F-MNIST for different heterogeneity scenarios, $\rho=1,2,5$. Variance is {depicted as a} 
    shaded area. 
    }
    \label{fig:acc_merged_scenarios}
\end{figure}

Different scenarios {exhibit varying energy profiles.}
Specifically, in the first scenario ({Figs.~\ref{fig:pick10users_energy}(a) and (d)}) where $\alpha = 1$ and {there is} only one partition ($\rho =1$), reaching higher accuracies {appears} 
to lead to an (almost) exponential growth {in} energy costs. 
We argue that this {phenomenon} is {closely} related to the model convergence. 
To 
better understand this relationship, Fig.~\ref{fig:acc_merged_scenarios} depicts the test accuracy as a function of the number of communication rounds. Notice that, when considering CIFAR-10 and {a single} partition,
{achieving test accuracies higher than $0.4$ requires fewer than $100$ communication rounds.}
{However, reaching an accuracy of $0.6$ requires almost all $500$ rounds to consistently achieve this level, resulting in more than tenfold higher energy consumption due to the slower convergence.}

A similar trend
is observed when comparing different levels of heterogeneity. For instance, while less than $17$Wh ($25\%$ of the relative energy) are {needed} 
to reach an accuracy of $0.45$ for $\rho = 1$ in Fig.~\ref{fig:pick10users_energy}(a), 
achieving an accuracy of $0.42$ with $\rho=5$ (Fig.~\ref{fig:pick10users_energy}(c)) requires almost $100$Wh  of energy (respectively, $142\%$ of the relative energy).
Once again, we argue that this is attributed to the quicker model convergence (i.e., less than $100$ communication rounds) to lower accuracies in scenarios with fewer partitions ($\rho = 1$).
Conversely, higher heterogeneity ($\rho=2$ and $\rho=5$) makes it more challenging for the model to reach the specified accuracy levels, thus leading to significantly higher energy consumption.
Consider, for example, the energy required to reach $0.4$ test accuracy in the CIFAR-10 dataset. This cost considerably escalates with an increasing number of partitions.
A similar behavior is observed {when aiming for} 
$0.75$ accuracy in the F-MNIST dataset\footnote{We note that
when considering the F-MNIST dataset and five partitions, better results could have been obtained by reducing the number of local epochs. However, we decided to perform $10$ rounds of local training to maintain consistency with the other cases. }.

\vspace{-1em}

\subsection{Overall Energy Consumption and Test Accuracy}

In this section, rather than comparing the different energy profiles of each client selection mechanism, we {conduct} 
a more in-depth analysis of the trade-off between total energy consumption, accuracy, and the number of communication rounds. 
Table~\ref{table:results_energy_acc} considers the same scenario as 
{previously discussed for} the CIFAR-10 dataset
but also presents the number of epochs needed to reach the desired accuracy level.
We also report the total (relative) energy and final test accuracy obtained {upon the completion of the FL training process.}
These results are presented in the last column 
of the table, {labeled} 
``Final Accuracy''.

\setlength{\tabcolsep}{5pt}
\renewcommand{\arraystretch}{1.25}
\begin{table*}[!ht]
\centering
\caption{
Energy 
{required} to consistently reach certain accuracy levels ($0.4$ and $0.5$) for at least $20$ consecutive rounds considering {the} CIFAR-10 dataset. 
The last column reports the (relative) total energy consumed during the FL process and the final test accuracy.
 Results are reported as the average of five different seeds. We also report the variance of all these quantities obtained from the different realizations. The reference energy is $70.10$Wh which is equivalent to $100\%$ of the relative energy. 
}
\label{table:results_energy_acc}
\begin{tabular}{l l cc cc cc}
        \toprule
        \rowcolor{rowgray}
 &                 & \multicolumn{2}{c}{\textbf{Accuracy 0.4}}     & \multicolumn{2}{c}{\textbf{Accuracy 0.5}}    & \multicolumn{2}{c}{\textbf{Final Accuracy}}          \\ 
 \rowcolor{rowgray}
\textbf{\begin{tabular}[c]{@{}c@{}}Scenario\\ ($\alpha$, $\rho$)-Dir\end{tabular}} & \textbf{Method} & \multicolumn{1}{c}{\textbf{\begin{tabular}[c]{@{}c@{}}Relative  \\ Energy (\%)\end{tabular}}} & \textbf{Rounds} & \multicolumn{1}{c}{\textbf{\begin{tabular}[c]{@{}c@{}}Relative \\ Energy (\%)\end{tabular}}} & \textbf{Rounds} & \multicolumn{1}{c}{\textbf{\begin{tabular}[c]{@{}c@{}}Relative \\ Energy (\%)\end{tabular}}} & \textbf{\begin{tabular}[c]{@{}c@{}}Test Accuracy\\ (\%)\end{tabular}}
\\ \midrule
        \multirow{4}{*}{(1,1)-Dir} 
       & {{Random}} 
& $18.41$  $(\pm 1.11)$ & $75.20$  $(\pm 1.64)$             & $30.38$  $(\pm 1.34)$ & $123.60$  $(\pm 6.58)$             & $123.85$  $(\pm 4.78)$ & $58.76$  $(\pm 0.90)$ \\

& {{FedCor}} 
& $25.68$  $(\pm 2.96)$ & $72.20$  $(\pm 4.66)$             & $38.22$  $(\pm 5.02)$ & $108.60$  $(\pm 10.88)$             & $171.79$  $(\pm 10.23)$ & $58.69$  $(\pm 0.92)$ \\

& {\textit{SimClust}} 
& $19.96$  $(\pm 2.09)$ & $79.20$  $(\pm 6.42)$             & $30.15$  $(\pm 3.52)$ & $119.00$  $(\pm 12.39)$             & $128.09$  $(\pm 4.27)$ & $58.66$  $(\pm 0.74)$ \\

& {\textit{RepClust}} 
& $17.82$  $(\pm 0.88)$ & $72.40$  $(\pm 4.22)$             & $26.50$  $(\pm 1.85)$ & $106.80$  $(\pm 7.53)$             & $125.66$  $(\pm 1.36)$ & $59.92$  $(\pm 0.43)$ \\      
        \midrule
        \multirow{4}{*}{(1,2)-Dir} 
& Random
& $71.44$  $(\pm 18.03)$ & $270.50$  $(\pm 66.58)$              & $-$  $-$ & $-$  $-$                   & $132.44$  $(\pm 2.78)$ & $45.16$  $(\pm 1.66)$ \\

& {FedCor}
& $63.15$  $(\pm 7.08)$ & $202.60$  $(\pm 23.64)$             & $-$  $-$ & $-$  $-$                  & $152.94$  $(\pm 1.66)$ & $48.08$  $(\pm 0.91)$ \\

& \textit{SimilarClust}
& $30.12$  $(\pm 4.40)$ & $114.25$  $(\pm 15.84)$             & $97.88$  $(\pm 21.77)$ & $368.75$  $(\pm 78.11)$             & $132.73$  $(\pm 1.76)$ & $51.63$  $(\pm 0.76)$ \\

& \textit{RepulsiveClust}
& $33.98$  $(\pm 3.03)$ & $129.00$  $(\pm 12.38)$             & $112.48$  $(\pm 18.08)$ & $419.50$  $(\pm 62.93)$             & $133.19$  $(\pm 1.40)$ & $49.90$  $(\pm 1.49)$ \\
        \midrule
        \multirow{4}{*}{(1,5)-Dir} 
        & {Random}
& $-$  $-$ & $-$  $-$             & $-$  $-$ & $-$  $-$             & $115.33$  $(\pm 1.21)$ & $35.59$  $(\pm 3.02)$ \\

& {FedCor}
& $113.68$  $(\pm 15.26)$ & $368.00$  $(\pm 48.84)$             & $-$  $-$ & $-$  $-$             & $153.81$  $(\pm 1.43)$ & $42.57$  $(\pm 1.36)$ \\

& \textit{SimClust}
& $93.20$  $(\pm 22.58)$ & $365.20$  $(\pm 80.36)$             & $-$  $-$ & $-$  $-$             & $127.26$  $(\pm 7.13)$ & $42.79$  $(\pm 2.16)$ \\

& \textit{RepClust}
& $86.75$  $(\pm 14.85)$ & $375.00$  $(\pm 63.47)$             & $-$  $-$ & $-$  $-$             & $115.74$  $(\pm 0.43)$ & $41.80$  $(\pm 1.64)$ \\
        \bottomrule
    \end{tabular}
\end{table*}

We start by noting that employing either of our proposed clustering mechanisms typically results in high test accuracies (comparable to or higher than \textit{FedCor}) while maintaining low energy consumption levels (similar to random selection).
This can be observed by comparing the ``Final Accuracy'' column 
of Table~\ref{table:results_energy_acc} for the different $(\alpha, \rho)$--Dir scenarios. 
Specifically, in the first scenario  ($(1,1)$--Dir), test accuracies are similar regardless of the method considered, with final accuracies around $0.58-0.59$ and within the variance of the other methods. 
However, employing active client selection (\textit{FedCor})  significantly increases the total energy costs, approximately $50$\% ($35.10$ Wh) with respect to the other methods. We argue that this increase can typically be associated {with} the ranking of clients and the training of the GP.

Furthermore, when considering the energy consumed to consistently reach a certain accuracy level, \textit{SimClust} and \textit{RepClust} typically require less amount of energy than  \textit{FedCor} and random selection. 
Even in scenarios where clustering methods {necessitate additional} 
communication rounds, the total energy consumption may still be lower or comparable to that of \textit{FedCor}. For instance, in the $(1,1)$--Dir setting with a desired accuracy of $0.4$, \textit{SimClust} requires more communication rounds than \textit{FedCor} but result in slightly lower energy consumption. A similar trend is observed when the desired accuracy is $0.5$.

We also observed similar energy profiles in more heterogeneous settings, with $\rho= 2$ and $\rho=5$. In these settings, however,  different client selection mechanisms strongly impact the final test accuracies and convergence rates.
Particularly, for $\rho=2$, clustering often leads to higher final test accuracies {compared to}
\textit{FedCor} or random client selection while consuming the least amount of energy. 
We argue that the accuracy gains obtained {with} 
the clustering solutions  (\textit{SimClust} or \textit{RepClust}) 
{stem}
from selecting clients that statistically approximate the information of the entire dataset, thereby potentially avoiding catastrophic forgetting.
This is not necessarily the case for \textit{FedCor} nor random selection, which (for a specific round) may select clients that, jointly, only contain information from one partition of the data. If this process is repeated over many rounds, it will inevitably lead to catastrophic forgetting. 
Note that when no partitions are considered (i.e.,  $\rho=1$), \textit{FedCor} typically requires fewer rounds than \textit{SimClust} and \textit{RepClust} to reach a desired accuracy. However, as 
the number of partitions {increases} ($\rho = 2$ or $\rho = 5$), the clustering solutions require the same or fewer communication rounds than \textit{FedCor} to reach different accuracy levels. 
Employing one of our clustering solutions (\textit{SimClust} or \textit{RepClust}) seems to always consume the 
{least} amount of energy during training. This {trend} is observed in both {the} early and final stages of the FL process, 
 {as indicated by} the different levels of test accuracy displayed in {Table~\ref{table:results_energy_acc}}. 

Finally, {when} comparing \textit{SimClust} and \textit{RepClust}, we empirically observed that the 
{former} often leads to slower convergence rates than the 
{latter} and requires more communication rounds to reach the same desired accuracy  (except for the scenario where $\rho=2$ in which case results could have been improved by using fewer local epochs).
Another {notable} difference is that the optimal number of clusters for \textit{SimClust} heavily {depends} 
on the existing label distribution. During our experiments, we noticed that running the same scenario, with fixed $(\alpha, \rho)$   but changing the seed, led to different optimal numbers of clusters.
This is not the case for \textit{RepClust}, where the optimal number of clusters remains relatively consistent across different executions of the same $(\alpha,\rho)$--Dir scenario. 

We argue that this is a consequence of how the groups are constructed in each of the clustering solutions. 
Specifically, clusters in \textit{RepClust} are {formed} with the objective that each cluster is similar to each other (e.g., similar label distribution, number of clients and samples), thereby minimizing the impact of selecting one cluster over another on the FL process. Conversely, \textit{SimClust} focuses on clustering similar clients together, which may lead to one cluster having a different label distribution and/or more clients and samples than another.

\subsection{Impact of Cluster Quality}

Up until now, when considering our clustering solutions, we have empirically selected the number of clusters $G$ that minimizes the total energy costs. In this final section, we analyze how the quality of the clustering impacts our proposed clustering-based solutions.

\begin{figure}[th]
    \centering
    \includegraphics[width=0.98\linewidth]{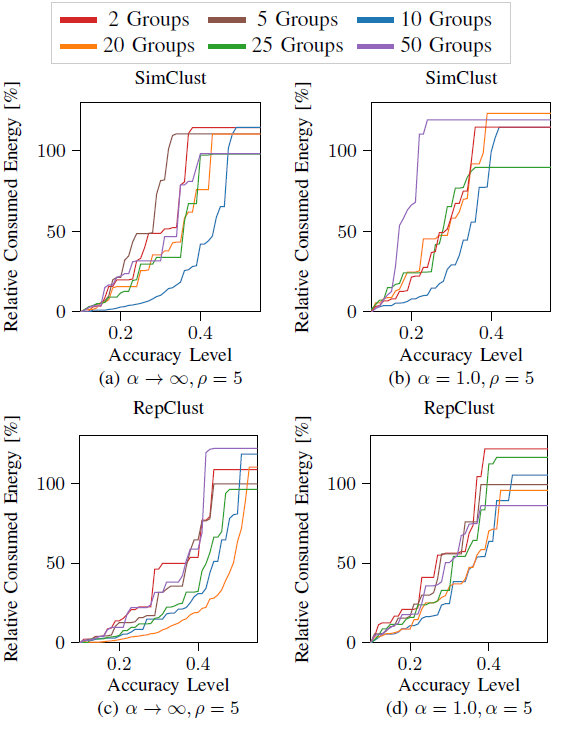}
    \caption{
    Impact in energy and accuracy of the number of clusters (different colors) for \textit{SimClust} and \textit{RepClust} in a homogeneous scenario with $\alpha \to \infty$, and a heterogeneous scenario with $\alpha = 1.0$. 
    }
    \label{fig:clustering_quality}
\end{figure}

We observed that varying $G$ and/or the number of elements in each cluster significantly influenced the total energy consumed.  
Fig.~\ref{fig:clustering_quality} presents the total energy necessary to consistently reach different accuracy levels on the CIFAR-10 dataset, for {various} numbers of groups ({represented by} different colors in the plots) and constant $K=10$. We consider the optimal number of clusters as the one that {achieves}
the highest accuracies while consuming the lowest amount of energy, i.e., the curves closest to the lower right corner of the plot. 
This time we slightly change the scenario discussed above and consider only one partition (i.e. $\rho = 1$) and vary $\alpha$ to simulate data heterogeneity. Specifically, we consider two scenarios: \textit{i}) a homogeneous scenario (dashed lines in the plot) where $\alpha \to \infty$ and \textit{ii}) a heterogeneous scenario in which $\alpha = 1.0$.

During our experiments, we observed that with \textit{SimClust}, the total energy consumed is strongly affected by the number of clusters, with similar numbers of clusters leading to different energy consumption levels. For instance, in Fig.~\ref{fig:clustering_quality}(a), the best results are obtained when considering $G=10$ groups, and choosing the wrong number of groups strongly affects the results. {Specifically,
slightly varying the number of clusters may lead to different results and the energy profile may depend on the heterogeneous scenario. }
In contrast, \textit{RepClust}  reaches better results when considering $20$ groups, but slightly varying the number of groups also results in similar energy profiles.  {Similarly, results obtained in the homogeneous scenario seem to be applicable also in the heterogeneous case.}
In this sense, \textit{RepClust} appears
to be more stable regardless of the number of clusters selected.

\subsection{Generalization of Proposed Approach}

Finally, we conduct a comprehensive set of experiments to compare our proposed approach against six FL sampling protocols across three benchmark datasets: F-MNIST, CIFAR-10, and CIFAR-100. Table~\ref{table:results_baselines} presents the test accuracy achieved by each of these methods under constrained energy budgets, specifically at $60\%$, $80\%$, and $100\%$ of the energy required for training with random sampling.
By analyzing accuracy under varying energy budgets, we can evaluate whether different methods experience significant performance degradation under constrained energy availability, e.g, in IoT devices running on battery. 
For reference, we recall that the energy consumption for full training (500 communication rounds and 10 epochs of local training) with random sampling is 80.58 Wh for F-MNIST, 70.10 Wh for CIFAR-10, and 658.22 Wh for CIFAR-100. 

Among the baselines, \textit{RepClust} achieves the highest accuracy for the F-MNIST dataset, while FLIS~\cite{morafah2023flis} performs best on the CIFAR-10 dataset.
However, FLIS also requires each device to share a portion of its local data to the central server, raising privacy concerns\footnote{Alternatively, the server may store synthetic data.}. 
Additionally, we conjecture that relying solely on the data transmitted by the devices may contribute to FLIS’s underperformance on CIFAR-100.  Specifically, it is well-known that ResNet18  exhibits a higher error rates and larger gradient magnitudes compared to deeper architectures (e.g., ResNet50). As a result, the central server, which relies on these gradients,  struggles to extract meaningful insights from the limited data it receives. This, in turn, affects both gradient approximations and the device sampling process, finally leading to degraded convergence and overall accuracy performance.

Among the baselines that do not require sharing local data with the server, \textit{FedCor} consistently outperforms the others both under energy constraints and in terms of final test accuracy—-except on CIFAR-100, where it performs worse than random sampling. Once again, we attribute performance issue to the high variability of gradients, which hinders \textit{FedCor}'s ability to effectively train the Gaussian process. Regarding the other two gradient-based methods, namely ClustLowVar~\cite{fraboni2021clustered} and DELTA~\cite{wang2024delta}, they generally lead to only marginal improvement over uniform random sampling when considering a limited energy budget. However, apart from the CIFAR-100 dataset, their final accuracy is typically higher than that of random sampling.

Overall, the results in Table~\ref{table:results_baselines} indicate that the majority of the baseline methods considered in this work exhibit strong dataset dependency, performing well on some datasets while underperforming on others. In contrast, our proposed clustering-based solutions demonstrate consistent performance under constrained energy budgets and in terms of final test accuracy across all datasets. Specifically, among the methods that do not rely on data sharing, \textit{RepClust} emerges as the best-performing solution, achieving the highest accuracy across all datasets.

\setlength{\tabcolsep}{9pt}
\renewcommand{\arraystretch}{1.25}
\begin{table*}[]
\centering
\caption{
Maximum test accuracy sustained for at least 20 consecutive rounds under energy constraints of $60\%, 80\%$ and $100\%$  of the reference energy—measured from a model trained with random sampling over $500$ rounds (F-MNIST: 80.58 Wh, CIFAR-10: 70.10 Wh, CIFAR-100: 658.22 Wh). The final column reports the final accuracy achieved by each method. The best results for each dataset and energy level are highlighted in bold.
}
\label{table:results_baselines}
\begin{tabular}{l l cccc}
    \toprule
    \rowcolor{rowgray} \textbf{Dataset} & \textbf{Method} & \textbf{60\% Energy} & \textbf{80\% Energy} & \textbf{100\% Energy} & \textbf{Final Acc} \\
    \midrule
\multirow{8}{*}{\textbf{F-MNIST}}        
& ClustLowVar~\cite{fraboni2021clustered}	& $49.34 \pm 1.67$ 	& $50.26 \pm 1.86$ 	& $50.36 \pm 1.76$ 	& $51.46 \pm 2.84$ \\
& DELTA\cite{wang2024delta}	& $44.90 \pm 2.14$ 	& $47.00 \pm 2.97$ 	& $48.39 \pm 2.45$ 	& $50.32 \pm 1.08$ \\
& FLIS~\cite{morafah2023flis}	& $52.43 \pm 1.45$ 	& $53.01 \pm 1.89$ 	& $54.49 \pm 1.18$ 	& $54.49 \pm 1.18$ \\
& PowerD~\cite{cho2020client_power_choice}	& $46.04 \pm 2.17$ 	& $46.59 \pm 2.12$ 	& $47.20 \pm 1.43$ 	& $48.57 \pm 2.31$ \\
& {Random}	& $46.03 \pm 0.64$ 	& $47.57 \pm 2.20$ 	& $48.36 \pm 3.14$ 	& $48.36 \pm 3.14$ \\
& {FedCor}~\cite{tang2022fedcor}	& $52.57 \pm 1.89$ 	& $52.95 \pm 2.52$ 	& $52.95 \pm 2.52$ 	& $53.06 \pm 2.39$ \\
& \textit{SimClust}	& $52.20 \pm 6.22$ 	& $53.28 \pm 5.24$ 	& $53.28 \pm 5.24$ 	& $53.28 \pm 5.24$ \\
& \textit{RepClust}	& $\mathbf{54.14 \pm 4.22}$ 	& $\mathbf{55.68 \pm 1.70}$ 	& $\mathbf{55.68 \pm 1.70}$ 	& $\mathbf{55.68 \pm 1.70}$ \\
\hline
\multirow{8}{*}{\textbf{CIFAR-10}}        
& ClustLowVar~\cite{fraboni2021clustered}	& $24.52 \pm 2.44$ 	& $29.89 \pm 3.24$ 	& $34.52 \pm 2.28$ 	& $38.78 \pm 1.28$ \\
& DELTA~\cite{wang2024delta}	& $21.37 \pm 1.90$ 	& $26.72 \pm 2.88$ 	& $29.23 \pm 2.66$ 	& $34.66 \pm 2.32$ \\
& FLIS~\cite{morafah2023flis}	& $\mathbf{39.13 \pm 2.25}$ 	& $\mathbf{41.86 \pm 1.24}$ 	& $\mathbf{43.48 \pm 0.87}$ 	& $\mathbf{47.00 \pm 0.92}$ \\
& PowerD~\cite{cho2020client_power_choice}	& $24.08 \pm 3.30$ 	& $26.38 \pm 3.07$ 	& $28.65 \pm 2.85$ 	& $34.95 \pm 2.37$ \\
& {Random}	& $24.37 \pm 3.06$ 	& $30.95 \pm 3.01$ 	& $33.36 \pm 1.71$ 	& $35.59 \pm 3.02$ \\
& {FedCor}~\cite{tang2022fedcor}	& $30.44 \pm 2.16$ 	& $33.99 \pm 0.76$ 	& $39.09 \pm 2.79$ 	& $42.57 \pm 1.36$ \\
& \textit{SimClust}	& $33.33 \pm 1.69$ 	& $37.65 \pm 3.24$ 	& $39.43 \pm 2.44$ 	& $42.79 \pm 2.16$ \\
& \textit{RepClust}	& $34.51 \pm 3.28$ 	& $37.79 \pm 2.27$ 	& $40.64 \pm 2.37$ 	& $41.80 \pm 1.64$ \\
%
%
\hline
\multirow{8}{*}{\textbf{CIFAR-100}}        
& ClustLowVar~\cite{fraboni2021clustered}	& $19.84 \pm 1.15$ 	& $20.96 \pm 0.72$ 	& $22.24 \pm 0.93$ 	& $25.77 \pm 0.45$ \\
& DELTA~\cite{wang2024delta}	& $21.56 \pm 0.65$ 	& $22.88 \pm 0.38$ 	& $24.18 \pm 0.37$ 	& $25.59 \pm 0.31$ \\
& FLIS~\cite{morafah2023flis}	& $23.68 \pm 0.63$ 	& $24.34 \pm 0.48$ 	& $25.12 \pm 1.00$ 	& $25.69 \pm 0.48$ \\
& PowerD~\cite{cho2020client_power_choice}	& $24.14 \pm 0.14$ 	& $24.74 \pm 0.46$ 	& $25.48 \pm 0.65$ 	& $25.48 \pm 0.65$ \\
& {Random}	& $24.43 \pm 0.73$ 	& $25.08 \pm 0.74$ 	& $25.85 \pm 0.66$ 	& $26.04 \pm 0.31$ \\
&  {FedCor}~\cite{tang2022fedcor}	& $23.75 \pm 0.41$ 	& $24.45 \pm 0.14$ 	& $24.88 \pm 0.44$ 	& $25.32 \pm 0.66$ \\
& \textit{SimClust}	& $24.20 \pm 0.59$ 	& $24.81 \pm 0.86$ 	& $25.47 \pm 0.54$ 	& $25.47 \pm 0.54$ \\
& \textit{RepClust}	& $\mathbf{25.33 \pm 0.60}$ 	& $\mathbf{26.07 \pm 0.62}$ 	& $\mathbf{26.32 \pm 0.63}$ 	& $\mathbf{26.32 \pm 0.63}$ \\
\hline
\end{tabular}
\end{table*}


\section{Conclusion}

In this paper, the total energy consumption of the FL process has been comprehensively studied, focusing on the energy expended during training, communication, and pre-processing for various client selection mechanisms. Addressing energy consumption is crucial not only for improving the sustainability of FL systems but also for making them more practical for deployment in resource-constrained environments{, such as those encountered in AIoT.} Specifically,  we observed that a significant portion of the total energy is typically consumed during local training. Our findings underscore the critical impact of client selection and clustering strategies on the energy efficiency and convergence aspects of the FL process. 
In particular, we noticed that performing random client selection may result in low pre-processing energy consumption but also low convergence rates. Conversely, employing an active client selection method such as \textit{FedCor} may expedite convergence at the cost of a more energy-intensive pre-processing phase. To address this, we propose two clustering-based client selection mechanisms, namely \textit{SimClust} and \textit{RepClust}, which we have empirically demonstrated to be efficient pre-processing solutions that accelerate convergence.

In the path forward, we highlight that our methods can be combined with techniques like over-the-air computing, quantization, and network pruning to further reduce energy consumption and the carbon footprint on the server side. Additionally, exploring clustering in a decentralized setup presents an intriguing direction, introducing new challenges such as data sharing and privacy. 
Integrating all these methods, while challenging, may lead to promising solutions for energy-efficient distributed learning {in AIoT systems}.

\appendices

\section{Experiment details} 
\label{sec:nn_details}

We employ two different neural network architectures depending on the dataset. For the CIFAR-10 experiments, we consider a CNN comprising 
three 
convolutional layers, each with a kernel size 
$3\times 3$ 
and
{output} 
channel sizes of $\{32, 64, 64\}$, respectively. 
Each convolutional layer is followed by $2\times2$ max pooling and a ReLu activation function. After the convolutional layers, there is a fully connected layer with $64$ neurons and a ReLu activation function, followed by a final fully connected classification layer with $10$ neurons
and a softmax activation function.  This combination of convolutional and fully connected layers results in a neural network with $73,418$ learnable parameters. 
In the experiments with the F-MNIST dataset, we employ an MLP with three fully connected layers of size $\{64, 30, 10\}$, respectively. The first two layers have a dropout of $0.5$ and a ReLu activation function whereas the last (classification) layer uses a softmax activation function. This MLP network has $29,034$ learnable parameters.

Regarding the FL process, we train both networks (CNN and MLP) using the stochastic gradient descent optimizer with a learning rate of $0.01$ and momentum of $0.5$.  Moreover, we perform $10$ epochs of local training with a batch size of $64$. {We consider a 70\%/30\% train/test data split ratio, which is randomly selected for every seed initialization. }
Finally, when considering active client selection using \textit{FedCor}, we adhere to the implementation and parameter settings outlined in~\cite{tang2022fedcor}. Specifically, we perform a global model warm-up for $10$ communication rounds, during which random selection is employed.
We train the GP with a maximum likelihood evaluation and set the annealing coefficient ($\beta$) to $0.9$. Similarly as~\cite{tang2022fedcor}, we utilize a linear kernel function of order one.

\section{ Scalability of Clustering Solutions: Numerical Analysis} 
\label{appendix:scalability_clustering}

\begin{figure}[!hb]
\centering
\includegraphics[width=0.98\linewidth]{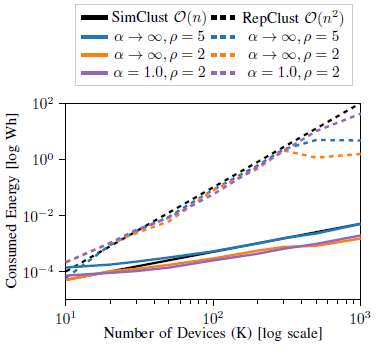}

    \caption{
    Energy consumed to perform clustering on CIFAR-10 under different data regimes (colors) using SimClust (solid) and RepClust (dashed). Values are in log-scale and  black lines illustrate time complexity trends. 
    }
    \label{fig:energy_vs_nb_devices}
\end{figure}

To corroborate with the theoretical time complexity described in Sec.~\ref{sec:scalability} and to analyze its relationship with energy consumption, we also conduct a set of numerical experiments using the CIFAR-10 dataset--the energy cost for clustering will be similar for F-MNIST as it is independent of the image content. Without loss of generality, we consider the scenario where the number of clusters $G$ is proportional to devices $L$ and partitions $\rho$. Specifically, for repulsive clustering, we set $G = L / \rho$ and for similarity clustering we use $G = \rho$. Fig.~\ref{fig:energy_vs_nb_devices} illustrates the energy required to perform clustering under different data regimes (indicated by different colors) for both similarity-based (solid lines) and repulsive (dashed lines) clustering. The values displayed were obtained as an average of 100 different seeds. To facilitate comparison, we also include trend lines (indicated in black) corresponding to $\mathcal{O}(n)$ and $\mathcal{O}(n^2)$. Moreover, for better visualization, results are presented on a logarithmic scale, where  $\mathcal{O}(n^2)$ grows twice faster than  $\mathcal{O}(n)$. 

To put these results into perspective, we start by recalling some numerical results presented in the beginning of Sec.~\ref{sec:results}, where training FedAvg with 10 clients randomly selected consumes approximately 70.10 Wh. The energy consumed during training is at least two to eight orders of magnitude higher than the energy required for clustering, as shown in Fig.~\ref{fig:energy_vs_nb_devices}. Moreover, because clustering only needs to be performed once before the training, its associated energy costs become negligible when compared to the learning phase.

\section{Differential Privacy: Clustering Noisy Labels}
\label{appendix:clustering_noise_label}

As mentioned above, directly sharing the label distribution from users to the server raises privacy concerns. In this appendix, we investigate the effect of applying differential privacy (DP) at device-level before transmitting the  label distributions to the server. The formal definition of differential privacy was first introduced in \cite{dwork2006differential} and further expanded in \cite{dwork2014algorithmic_book}. Since then, the method has been widely applied in ML domain (see \cite{wei2020federated} and references therein).

Majority of FL-based DP methods focus on gradient perturbation \cite{wei2020federated_dp} as the goal is to ensure privacy over the local updates. However, since we aim to protect the label distribution itself, we adopt a different strategy and directly add noise to the label distribution, at the local device, before sending it to the server.
Specifically, we consider the Gaussian differential privacy mechanism, a standard approach which perturbs the original  distribution as 
\begin{equation}
    \tilde{p}_{j,m} = p_{j,m} + \varepsilon_\sigma, \qquad j = 1,\ldots, L
\end{equation}
where $p_{j,m}$ represents the relative (ranging from $0-1$) label distribution associated to the $m$th class at the $j$th device, and $\varepsilon_\sigma \sim \mathcal{N}(0, \sigma^2)$ is a Gaussian random variable with variance $\sigma^2$. To ensure privacy while preserving the statistical properties of the data, we set
$$
\sigma = \frac{\gamma}{M L}\sum_j^L ||p_j||_1, \qquad j = 1,\ldots, L
$$
with $\gamma \geq 0$ controls the privacy level, $M$ is the total number of classes in the dataset and $L$ the number of devices in the system. 
This simple solution masks the device's label distribution while preserving the overall statistical properties of the label distributions, thus providing differential privacy guarantees. Introducing the variable $\gamma$ allows us to control the level of privacy (increasing $\gamma$ translates to higher privacy) regardless of the dataset or label distribution.

\begin{figure}[t!]
\centering
\includegraphics[width=0.98\linewidth]{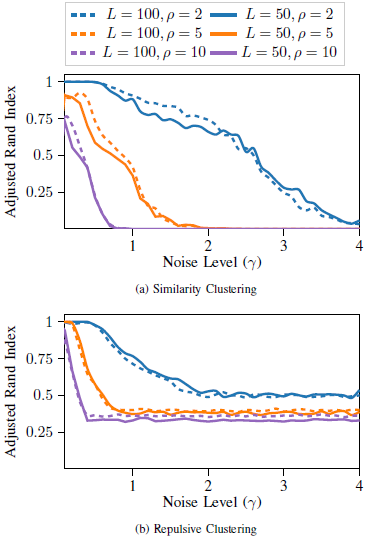}

         \caption{Accuracy plots for (a) CIFAR and (b) F-MNIST for different heterogeneity scenarios, $\rho=1,2,5$. Variance is {depicted as a} 
    shaded area. 
    }
    \label{fig:appendix:acc_merged_scenarios}
    \vspace{-1\baselineskip}
\end{figure}

To assess our proposed clustering approach under privacy constraints, we measure the quality of the predicted clustering against the true solution using the adjusted rand index (ARI), which requires knowledge of the true clustering solution. 
Consequently, we consider the scenario where devices have homogeneous data distribution  (i.e., $\alpha \to \infty$), which allows for an accurate estimation of the true clustering solution\footnote{For heterogeneous settings (different values of $\alpha$) or $\rho = 1$, estimating the true clustering solution is an NP-hard problem that requires trying out all possible solutions.}. ARI values close to 1 indicate perfect agreement between the predicted clustering solution and the true one, whereas values close to 0 suggest that the predicted solution is as good as random assignment.  Fig.~\ref{fig:appendix:acc_merged_scenarios} displays the ARI (y-axis) for increasing privacy levels $\gamma$ (x-axis) and different settings (colors) using CIFAR-10 dataset. From these results, we draw three key observations regarding privacy and scalability.

Regarding privacy, it is expected that increasing the noise level--and consequently enhancing privacy--will degrade clustering performance. This trend is consistently observed in both SimClust and RepClust, highlighting the inherent trade-off between privacy and clustering accuracy. However,  a different pattern is observed in each method. Specifically, SimClust (Fig.~\ref{fig:appendix:acc_merged_scenarios}(a)), high levels of noise eventually lead the ARI to zero, indicating that the clustering solutions are similar to random choice. Conversely, RepClust (Fig.~\ref{fig:appendix:acc_merged_scenarios}(b)) tends to an ARI value higher than zero, suggesting that stratified clustering is more robust to noise than performing similarity-based clustering. We argue that this trend may stem from the fact that it is easier to group together diverse clients, each containing data from multiple partitions, than to correctly cluster clients belonging to a single partition when noise is added to the true label distribution.

Finally, in addition to the scalability results discussed in Appendix~\ref{appendix:scalability_clustering}, Fig.~\ref{fig:appendix:acc_merged_scenarios} also shows that increasing the number of devices from $50$ (solid lines) to $100$ (dashed lines) has minimal impact on clustering quality. 
This is expected, as clustering is performed on relative label distributions rather than absolute samples-to-class counts. Consequently, scaling the number of clients while preserving the proportional distribution of labels results in similar clustering performance. 
On the other hand, in the more complex setting, increasing the number of partitions $\rho$ (indicated by different colors) consistently degrades performance, highlighting that greater data heterogeneity makes clustering more challenging.

\bibliographystyle{IEEEtran}
\bibliography{./bibliography.bib}

\end{document}